\documentclass[12pt, a4paper, twocolumn ]{article}
\usepackage{amssymb,amsmath}
\usepackage{graphicx}
\usepackage{hyperref}

 \renewcommand\Re{\operatorname{\mathfrak{Re}}}

 \DeclareMathOperator{\tr}{Tr}

\begin{document}
\title{Challenge IEEE-ISBI/TCB :\\ \large{Application of Covariance matrices and wavelet marginals}}
\author{Florian Yger \\ Tokyo Institute of Technology - \texttt{florian@sg.cs.titech.ac.jp} 
}
\date{\today}
\maketitle
\abstract{This short memo aims at explaining our approach for the challenge IEEE-ISBI on Bone Texture Characterization. In this work, we focus on the use of covariance matrices and wavelet marginals in an SVM classifier.}

\section{Introduction}
\begin{figure}[ht]
\centering
\includegraphics[width=.4\columnwidth]{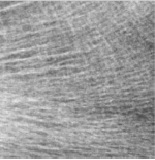}\quad
\includegraphics[width=.4\columnwidth]{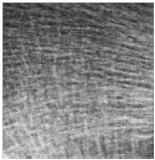}
\caption{Example of textures of a control subject (left) and a patient  with osteoporotic fractures(right).}
\label{fig:BoneTexture}
\end{figure}

Texture Characterization of Bone radiograph images (TCB) is a challenge in the osteoporosis diagnosis organized for the International Society for Biomedical Imaging (ISBI) 2014. The goal of this Challenge is to identify osteoporotic cases from healthy controls on 2D bone radiograph images, using texture analysis. The dataset consists of two populations composed of 87 control subjects (CT, Figure~\ref{fig:BoneTexture} (left)) and 87 patients with osteoporotic fractures (OP, Figure~\ref{fig:BoneTexture} (right)).

As illustrated by Figure~\ref{fig:BoneTexture}, textured images from the bone microarchitecture of osteoporotic and healthy subjects are very similar, making the challenge's task highly difficult .

\section{Feature for textures}
In our submissions to the ISBI challenge on texture classification, we have not looked for complicated application specific features or for a fancy feature selection algorithm. We rather focusd on two simple types of features, namely covariance matrices and wavelet marginals. Those submissions aimed at evaluation the features already studied to a real-life application. 

\subsection{Covariance matrices}
Covariance matrices have been studied as image descriptor in wide variety of applications from licence plate detection~\cite{porikli2006robust} to pedestrian detection~\cite{tuzel2007human}.

For an image or a region of an image $I \in \mathbb{R}^{d_1 \times d_2}$, this approach consist in computing local features $f(x)$ (usually statistical properties) on every pixel $p_{ij}$. Then, for of those local descriptors, the unbiased  empirical estimator of the covariance matrix  is computed as :
\begin{equation}\label{eq:covMat}
C=\frac{1}{n-1}\sum_{i, j} (f_{ij} - \bar{f})(f_{ij} - \bar{f})^\top
\end{equation}
with $n=d_1 \times d_2$ and $\bar{f}$ being the empirical mean of $f$. Note that this estimator will be accurate provided that the number of samples is large enough compared to the number of features.

However, this estimator is well-known for its sensitivity to outliers. To overcome this issue, a robust estimator -Minimumum Covariance Determinant (MCD)- has been introduced in~\cite{rousseeuw1984least}. Basically, MCD aims at finding $h$ observations (out of $n$) whose covariance matrix has the lowest determinant. 
\\Even if it suffered recently some controversies about its convergence properties, we use the algorithm (FastMCD~\cite{rousseeuw1999fast}) that has been proposed to approximate the MCD estimator. We use the implementation provided in the LIBRA toolbox~\footnote{Available at \url{http://wis.kuleuven.be/stat/robust/LIBRA/LIBRA-home}.} for MATLAB~\cite{verboven2005libra}. In our experiment, when using the FastMCD algorithm, we set $\alpha=0.9$ (meaning that the algorithm should be robust up to $10\%$ of outliers) 
and $n_{\text{trial}} = 500$ the number of trial subsamples drawn from the dataset.

Concerning the local features used for computing a covariance matrix, there exists several choices. We used two variants of features used in the litterature :
\begin{itemize}
\item gradient based~\cite{porikli2006robust,tuzel2007human} : 
\begin{align}
f_{ij} &= \Bigl[I^{ij}, |I_x^{ij}|, |I_y^{ij}|,|I_{xx}^{ij}|, |I_{yy}^{ij}|,\ldots \\ \notag
\ldots & \sqrt{(I_x^{ij})^2+ (I_y^{ij})^2}, \arctan \frac{|I_x^{ij}|}{|I_y^{ij}|} \Bigr]^\top
\end{align} 
where $I^{ij}$ is the intensity of the pixel $(i,j)$ and $I_x$, $I_{xx}$,.. are the intensity
derivatives (first and second order along the $x$ and $y$ axis) and the last term is the edge orientation
\footnote{Note that contrary to cited paper, we did not use the pixel coordinates as it did not make sense for texture analysis and gave poor results.}, leading to a ${7 \times 7}$ covariance matrix.
\item Gabor based~\cite{tou2009gabor,palaio2009kernel} :
\begin{align}
f_{ij} &= \Bigl[g_1^{ij}(I), \ldots g_p^{ij}(I) \Bigr]^\top
\end{align}  
where we have 
${g_1(I)=\sqrt{ ||\Re (I \star g_1)||^2 }}$, norm of the real part of the convolution of the image $I$ with a Gabor filter $g_1$. In our experiments, we used a filter bank of Gabor filters with parameters $\gamma =1$, $\theta \in \{ -\frac{\pi}{4}, 0 ,\frac{\pi}{4}, \frac{\pi}{2}\}$ and $\sigma_x = \sigma_y \in \{ 5,10, 20\}$, leading to a ${12 \times 12}$ covariance matrix.
\end{itemize}

As already noticed in the litterature~\cite{Tuzel06regioncovariance,porikli2006robust, porikli06covariancetracking,tuzel2007human}, covariance matrices belong to a non-Euclidean space where distances are not computed on straight lines but rather on curves lines (namely geodesics). Hence, using tools from the Riemannian geometry to analysis, it is possible to analyse those structured data. For example, given  $C_i$ and $C_j$ two non-singular covariance matrices, the Riemannian distance between them is :
\begin{equation}\label{eq:distRiem}
\delta_R (C_i, C_i)=||\log (C_i^{-\frac{1}{2}} C_j C_i^{-\frac{1}{2}})||_{\mathcal{F}}
\end{equation}
with $||.||_{\mathcal{F}}$ the Frobenius norm and $\log(.)$ the matrix principal logarithm.

\begin{figure}[t]
\includegraphics[width=0.95\columnwidth]{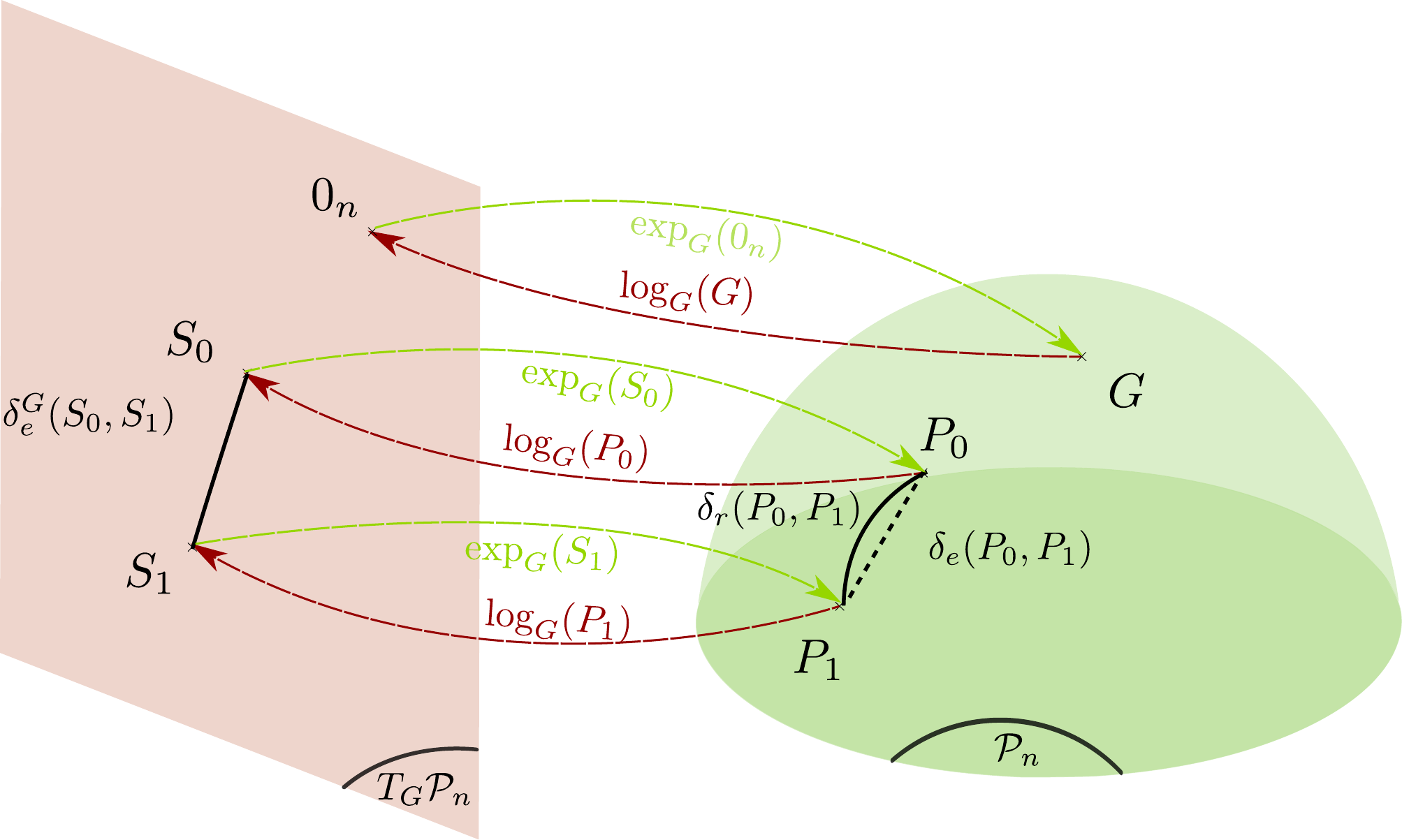} 
    \caption{Mapping between the Riemannian manifold of Symmetric Definite matrices and its Tangent space at the identity. (extracted from~\cite{yger2013review})}
    \label{fig:mappingLogEuclidean}
\end{figure}

\begin{figure}[t]
\includegraphics[width=0.9\columnwidth]{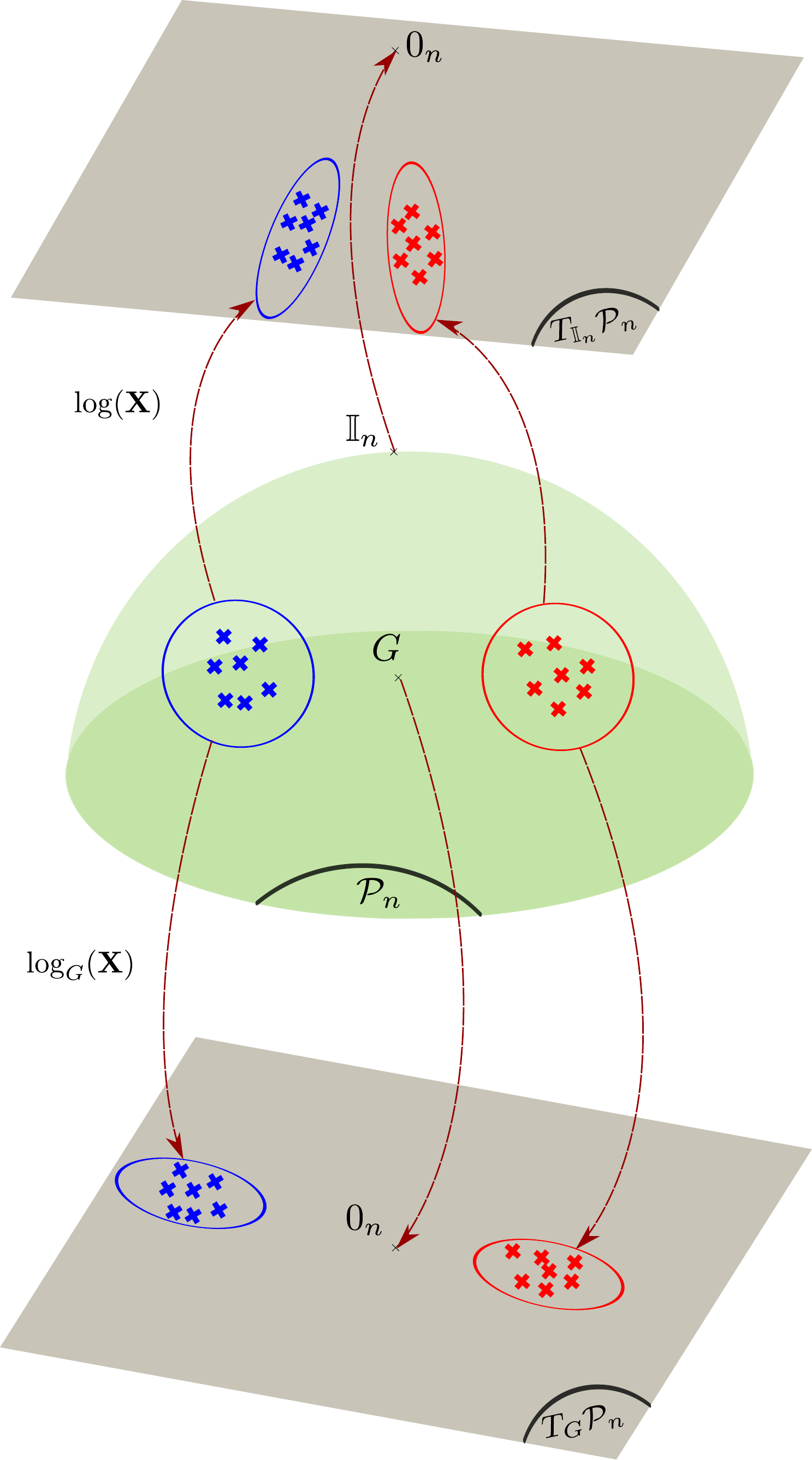} 
    \caption{Illustration of the deformation induced by the choice of different reference of tangent space. (extracted from~\cite{yger2013review})}
    \label{fig:LogEuclideanGeom}
\end{figure}

Recently, some authors investigated the use of such a feature for EEG signals and propose to use different kernels of the litterature to handle it~\cite{barachant2013classification,yger2013review}. We intend to apply those study to covariance matrices computed on images. 
\\In our experiments on textures (and coherently to the results in~\cite{yger2013review}), normalized LogEuclidean kernels showed the best performance in a Leave-one-out cross-validation. For two non-singular covariance matrices, this kernel $k_G (C_i, C_j)$ is defined as~:
\begin{align}
\label{eq:LogEuclidean}
\frac{\tr (\log (G^{-\frac{1}{2}} C_i G^{-\frac{1}{2}})\log (G^{-\frac{1}{2}} C_j G^{-\frac{1}{2}}))}{||\log (G^{-\frac{1}{2}} C_i G^{-\frac{1}{2}})||_{\mathcal{F}}||\log (G^{-\frac{1}{2}} C_j G^{-\frac{1}{2}})||_{\mathcal{F}}}
\end{align}
with the parameter $G$ being a non-singular covariance matrix. This kernel can be understood as a normalized scalar product in a the Tangent space\footnote{It is a local linear approximation around $G$ of the Riemannian manifold.} around $G$ (a Euclidean space where the data are mapped by the logarithm application - see Figure~\ref{fig:mappingLogEuclidean}) to the space of positive definite matrices. As the property demonstrated in~\cite{yger2013review} suggest it, the choice of  $G$ induces a deformation of the shapes in the feature space. So far, two heuristics have been used, either the identity matrix $\mathbb{I}$ or the Riemannian mean~\cite{moakher2005differential} of the learning set $G_{rm}$.

This deformation of the geometry induced by the use of a LogEuclidean kernel is illustrated in Figure~\ref{fig:LogEuclideanGeom}. Hence, it seems reasonnable to use the Riemannian mean as it may reduce the amount of distortion induced by flattening the manifold.

\subsection{Wavelet marginals}
Wavelet marginals are signal and image descriptors based on wavelet decomposition. This feature has been developed in order to extract frequential information for translation invariant classification of biomedical signals~\cite{farina2007optimization} and textures~\cite{yger2011wavelet}. 

Before delving into the description of this feature, let us briefly introduce some notations in the context of one-dimensional signals\footnote{For a comprehensive review of wavelet decomposition, the reader should refer to~\cite{mallat2008wavelet}.}.
Let $\phi_{\theta}$ be a mother wavelet (which shape is parametrized by $\theta$). We denote by $\theta_{\theta,s,t}$ the wavelet obtained from the mother wavelet after a dilatation at scale $s$ and a translation $t$.

As originally described in~\cite{farina2007optimization}, it is possible to extract the information contained in some frequency bands using wavelet marginals. For a signal $x \in \mathbb{R}^d$, for every $s \in [0,\cdots,\overbrace{\log_2(d)}^{J}]$, this feature is defined as :
\begin{equation}\label{eq:marginalLucas}
m_{\theta,s}(x)= \frac{\sum_t |\langle \phi_{\theta,s,t},x \rangle|}{\sum_t\sum_s |\langle \phi_{\theta,s,t},x \rangle|}
\end{equation}

\begin{figure}[t]
\includegraphics[width=0.95\columnwidth]{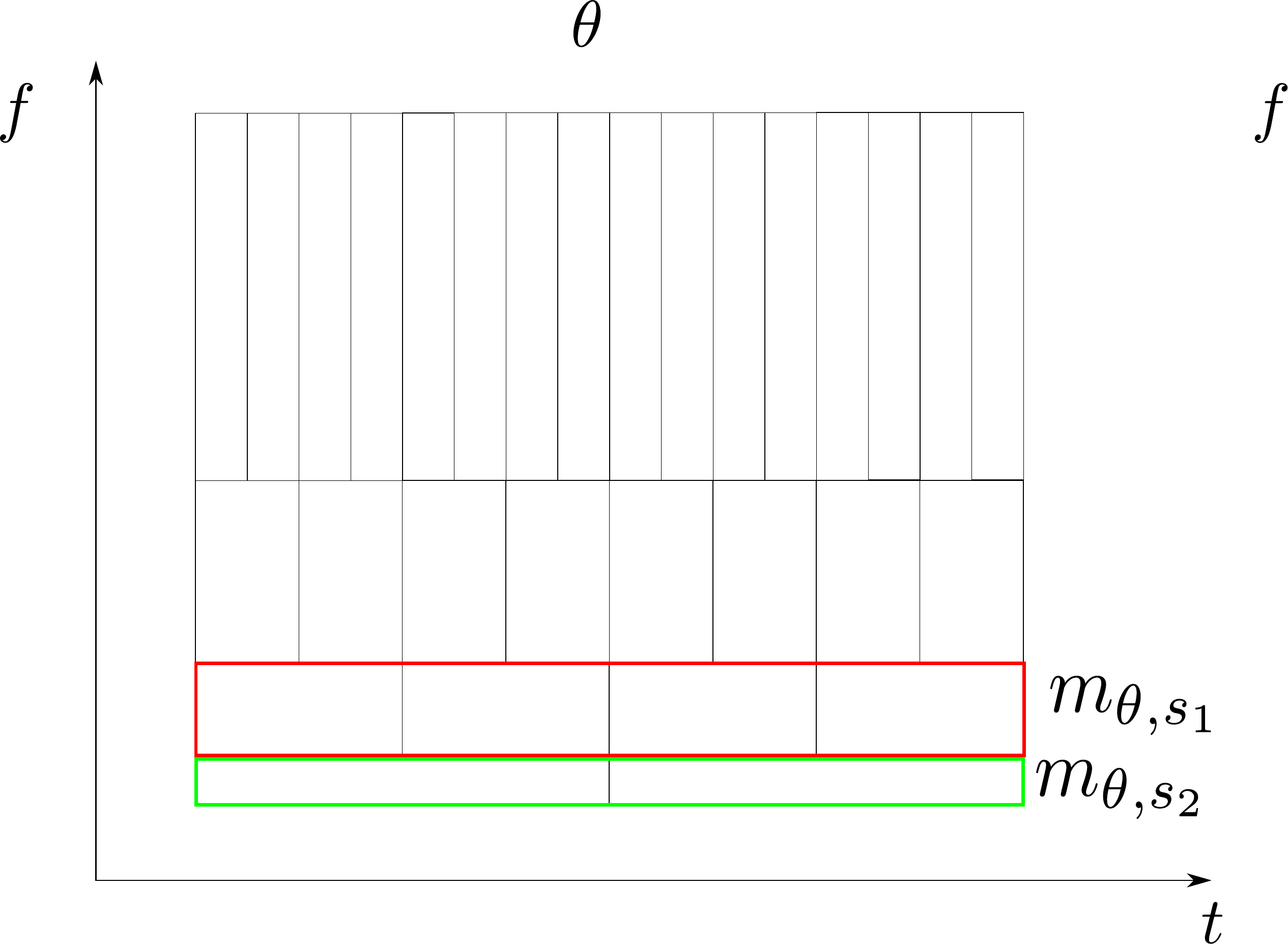} 
    \caption{Illustration of the information extracted in the time-scale space by the two marginals (at scale $s_1$ and $s_2$) of a wavelet decomposition (with a waveform parametrized by $\theta$. }
    \label{fig:marginal}
\end{figure}

For a given signal of size $d$, it is possible to extract at most $\log_2(d)$ marginals. As illustrated in Figure~\ref{fig:marginal}, every marginal extract this information contained in a given frequency band of the signals.

Once the wavelet decomposition of a two-dimensional image is defined, it is straightforward to extend marginals to images. Let $\psi_theta$ be a scaling function and $\phi_theta$ its corresponding mother wavelet. For the purpose of image analysis~\cite{mallat2008wavelet}, three different 2D mother wavelets are generated from the tensor product of the wavelet and the scaling function. Then, with $t_x$ and $t_y$ the translations along x and y axis respectively, we have the following wavelet coefficients for an image $I \in \mathbb{R}^{d \times d}$ :
  \begin{equation}\label{eq:wavelet2d}
c_{\theta,s,t_x, t_y}^l (I)=\langle \phi_{\theta,s,t_x, t_y}^l, I \rangle
\text{~} l=\{1,2,3\}
\end{equation}

As proposed in~\cite{yger2011wavelet}, by summing over the extra indices ($l$ and $t_y$) in Eq.\ref{eq:marginalLucas}, it is possible to extend this feature to images.

In this work, marginals are used as a baseline. Indeed, this feature is very sensitive to the waveform used for analysis the signal. Some method has been proposed~\cite{yger2011wavelet} in order to optimize this waveform and to select the relevant scale for classification.\\
For a given image, we decomposed it using a Haar wavelet\footnote{The wavelet decomposition was performed using Wavelab 850 toolbox for Matlab available at : \url{http://statweb.stanford.edu/~wavelab} } and then computed the marginals of the decomposition for every scale. Using the labeled dataset, we normalize the data to a zero mean and unit variance and then a linear kernel was used.  Note that since unnormalized marginals are positive and sum to one, the use of $\chi^2$ kernels\cite{haasdonk2004learning} may be investigated.

\section{Methodology}

\subsection{Image preprocessing}
For extracting wavelet marginals, we need the images to have dyadic dimensions. Hence, we resized (using the Matlab function \textit{imresize}) the image from $400 \times 400$ to $256\times 256 $, $128 \times 128$ or $64 \times 64$.
\\Finally, based on our validation results, marginals of Haar wavelets seemed to be the most efficient on $128 \times 128$ images.

On our first round of experiments, we applied gradient based covariance matrices to the raw images and obtained very low (almost random) validation results. The gradient based features being very localised, it seemed to miss some important information on the data. Hence, we resized the images by different factors in order to extract more global inforamtion. We rescaled the images to factor $\frac{1}{2}$,$\frac{1}{4}$,$\frac{1}{8}$ or $\frac{1}{16}$. On our validation procedure, a rescaling factor of $\frac{1}{8}$ (e.g. $50 \times 50$) gave the best validation performance.
\\However, when computing our covariance matrices, we observed very unstable results. As this estimator is not robust to outliers, we applied the FastMCD algorithm to approximate the MCD estimator. For gradient based features, we obtained a boost in the validation results (compared to the empirical covariance matrices).

For the Gabor based covariance matrices, the results have been somehow very different. We observed that applying Gabor based covariance matrices to rescaled images was giving worst validation results. This may be because the Gabor filters used were already extracting global information on the raw images. The choices of the parameters of the Gabor filters considered have been choosen based on their validation performance.
\\Note that contrary to the gradient based features, we did not use the FastCMD algorithm and only relied on the empirical covariance matrices of the Gabor features. Indeed, the FastCMD approximation lead to poor validation results. However, this may only indicate that we should have bette tune the parameter of the FastCMD algorithm (and raising the number trial $n_{\text{trial}}$.

\subsection{Validation procedure}
\begin{table*}[t]
  \centering
  \begin{tabular}{|l||c|c|c|}
  \hline
  feature & CmdMat-grad & CovMat-gab & Marginal-Haar\\
  image size & $50\times 50$ & $400 \times 400$ & $128 \times 128$ \\\hline
  kernel type & LogEuclidean & LogEuclidean & linear \\
  kernel parameter & identity& Riemannian mean & - \\\hline
  validation & $83.22\%$& $90.63\%$ & $64.26\%$\\
  LOO accuracy & $77.59\%$ & $74.14\%$ & $60.34\%$\\ \hline
    
  \end{tabular}
  \caption{Properties and mean accuracy over the Leave-one-out cross-validation procedure for our methods.}
  \label{tab:LOOResults}
\end{table*}

The rules of the competition specified that the competitors had to use an SVM classifier\footnote{We used the SVM toolbox for MATLAB - available at \url{http://asi.insa-rouen.fr/enseignants/~arakoto/toolbox/index.html}. }. We tuned the hyperparameter $C$ of the classifier using a Leave-one-out (LOO) cross-validation procedure. This parameter could take values in the set $\{1,10,100,1000,10000,100000\}$. 

When we produced two variants of the same methods (for example, two preprocessing different for the raw images or same feature with different kernels), we selected the variant that achieved the best mean accuracy over the LOO procedure. We also report as \emph{validation}, the mean accuracy of the learned classifiers on the training dataset.

We sum up the obtained results in Tab~\ref{tab:LOOResults} and the properties of the proposed approach. 

\section{Conclusion and perspectives}

Based on the validation results, it was difficult to choose one method from the three proposed. Indeed, the big gap of accuracy between the LOO and validation results led us to fear for overfitting. On the other hand, despite worst results, using the \emph{Marginal-Haar} features shows closer validation and LOO accuracy criterion.

After having consulted the competition organizers, we submitted the three methods and obtained surprizing results.

\subsection{Competition results}
We first report our final results as announced by the organizers~\footnote{The final results are available at \url{http://www.univ-orleans.fr/i3mto/results}}

In this Tab~\ref{tab:Results}, we report the criteria used by the competition organizers :

\begin{description}
\item{TP - True Positive :} number of subjects with Osteoporosis correctly identified
\item{FP -False Positive :} number of Control subjects incorrectly identified
\item{TN - True Negative :} number of Control subjects correctly identified
\item{FN - False Negative :} number of subjects with Osteoporosis incorrectly identified
\item{Sn - Sensitivity :} defined as $Sn = \frac{TP}{TP + FN}$
\item{Sp - Specificity :} defined as $Sp = \frac{TN}{FP + TN}$
\end{description}

\begin{table*}[t]
  \centering
  \tabcolsep=0.16cm
  \begin{tabular}{|l||c|c|c|c|c|c||c|c|c|c|c|c||c|}
  \hline
  & \multicolumn{6}{c||}{First results} & \multicolumn{7}{c|}{Blind results} \\ \cline{2-14}
  Method & TP & FP & TN & FN & Sn & Sp &TP & FP & TN & FN & Sn & Sp & rank\\
  \hline
Marginal-Haar & 36  & 20 & 38 & 22 & 0.62 & 0.66 & 19 & 10 & 19 & 10 & 0.66 & 0.66 & 1 \\
CmdMat-grad & 54 & 7 & 51 & 4 & 0.93 & 0.88 & 16 & 15 & 14 & 13& 0.55 & 0.48& 5 \\
CovMat-gab & 46 & 7 & 51 & 12 & 0.79 & 0.88 & 13 & 14 & 15 & 16 & 0.45 & 0.52 & 6 \\ \hline
  \end{tabular}
  \caption{Published results on the TCB competition for the three proposed methods.}
  \label{tab:Results}
\end{table*}

From those criteria, the challenge organizers derived six other criteria used to rank the submitted methods. In Tab~\ref{tab:Results}, we report the mean rank on those $6$ criteria as communicated by the challenge organizers.

From the gap between the first and blind results, it clearly appears that both covariance based methods overfitted and obtained deceiving results.\\
It should be stated that the Mariginal based method has been ranked first on every of the $6$ criteria used by the organizers. 

\subsection{Perspectives}
As stated in the introduction, we have not applied state-of-the art features in texture classification but rather tried to apply previously proposed work. Indeed, it should be noted the recently proposed scattering transform~\cite{sifre2012combined,bruna2013invariant} may be a more powerful texture descriptor than what we proposed.

It should also be noted that Wavelet marginals have been used in a rather different setting than their original proposition~\cite{yger2011wavelet}. Indeeed, we restricted ourself to the use of a single Haar wavelet basis but since there is a strong impact on the choosen wavelet parametrizing a marginal, we should have validated carefully this choice. \\In the original method, the wavelet parameter was  selected through an MKL based approach that could not be applied for this competition since the rules restricted the use of SVM classifiers only (implicitly forbidding MKL methods). Moreover, as the final results suggest it, the main issue in this competition resides in overfitting, so an MKL approach having more degree of liberty, its should be very carefully tuned.

In the same line of thought, combining different features (through an MKL method) has shown very good practical results in~\cite{gehler2009feature}. For a real world application where enough data are available, this would be a very promising future work. Yet, in a context of data competition (with only limited data), such an MKL approach may lead to overfitting.

\section*{Acknowledgments}
Part of this work has been carried in the laboratory LITIS with the support of a grant from LeMOn ANR-11-JS02-10 and it was finished in Tokyo Institute of Technology with the support of a JSPS fellowship.

\bibliographystyle{plain}
 \bibliography{TBCchallenge}

\end{document}